 \documentclass[pmlr,twocolumn,10pt]{jmlr} 
 
 \let\SUP\textsuperscript

\usepackage{tabto}    
\usepackage{bm}
\usepackage{graphicx}
\graphicspath{ {images/} }
\usepackage{float}
\usepackage{natbib}
\usepackage{url}
\usepackage{textcomp}

\usepackage{booktabs}
\usepackage[load-configurations=version-1]{siunitx} 

\theorembodyfont{\upshape}
\theoremheaderfont{\scshape}
\theorempostheader{:}
\theoremsep{\newline}

\jmlrvolume{ML4H Extended Abstract Arxiv Index}
\jmlryear{2021}
\jmlrsubmitted{2021}
\jmlrpublished{}
\jmlrworkshop{Machine Learning for Health (ML4H) 2021} 

\title[NLP Techniques to Detect Cognitive Impairment]{Using Deep Learning to Identify Patients with Cognitive Impairment in Electronic Health Records}

\author{Tanish Tyagi\nametag{\thanks{Authors contributed equally}\SUP{1}},
Colin G. Magdamo\nametag{\footnotemark[1]\SUP{1}}, 
Ayush Noori\SUP{1},
Zhaozhi Li\SUP{1},
Xiao Liu\SUP{1},
Mayuresh Deodhar\SUP{1},
Zhuoqiao Hong\SUP{1},
Wendong Ge\SUP{1},
Elissa M. Ye\SUP{1},
Yi-han Sheu\SUP{1},
Haitham Alabsi\SUP{1},
Laura Brenner\SUP{1},
Gregory K. Robbins\SUP{1},
Sahar Zafar\SUP{1}, 
Nicole Benson\SUP{1},
Lidia Moura\SUP{1},
John Hsu\SUP{1},
Alberto Serrano-Pozo\SUP{1},
Dimitry Prokopenko\SUP{1},
Rudolph E. Tanzi\SUP{1},
Bradley T. Hyman\SUP{1},
Deborah Blacker\SUP{1},
Shibani S. Mukerji\SUP{1},
M. Brandon Westover\SUP{1},
Sudeshna Das\SUP{1}
\centering \Email{
\\[\bigskipamount] 
\SUP{1}\{ttyagi, 
cmagdamo,
anoori1,
zli39,
xliu61,
mdeodhar,
zhong1,
wendong.ge,
emye,
ysheu,
halabsi,
lnbrenner,
grobbins,
sfzafar, 
nbenson, 
lidia.moura, 
john.hsu,
aserrano1,
dprokopenko,
dblacker,
rtanzi,
bhyman,
smukerji,
mwestover,
sdas5\}
@mgh.harvard.edu}
\centering
\begin{center}\addr Massachusetts General Hospital and McCance Center for Brain Health, Boston, MA 
\end{center}
}

\begin{document}

\maketitle

\begin{abstract}
\hspace{10mm} Dementia is a neurodegenerative disorder that causes cognitive decline and affects more than 50 million people worldwide. Dementia is underdiagnosed by healthcare professionals — only one in four people who suffer from dementia are diagnosed. Even when a diagnosis is made, it may not be entered as a structured International Classification of Diseases (ICD) diagnosis code in a patient’s charts. Indeed, information relevant to cognitive impairment (CI) is often found within electronic health records (EHR) but manual review of clinician notes by experts is both time consuming and often prone to errors. Automated mining of these notes presents an opportunity to label patients with cognitive impairment in EHR data. We developed natural language processing (NLP) tools to identify patients with cognitive impairment and demonstrate that linguistic context enhances performance for the classification task. We fine-tuned our attention based deep learning model, which can learn from complex language structures, and substantially improved accuracy (0.93) relative to a baseline TF-IDF (term frequency-inverse document frequency) NLP model (0.84). Further, we show that deep learning NLP can successfully identify dementia patients without dementia-related ICD codes or medications.
\end{abstract}


\section{Introduction}
\label{sec:intro} Dementia is the most common neurodegenerative disease affecting older adults, progressing from mild cognitive impairment (MCI) to mild, moderate, and severe dementia. Dementia is underdiagnosed \citep{AlzheimerAssociation}: dementia is not formally diagnosed or coded in claims data for over 50\% of older adults living with probable dementia. Often, a diagnosis is given once patient has reached moderate dementia, and irreversible damage has already been done to the brain.  
The early detection of the first signs of cognitive impairment, however, is important for improving clinical outcomes and patient management. Tools that can efficiently and effectively analyze medical records for warning signs of dementia and recommend patients for follow up with a specialist can be critical to obtaining an early diagnosis for dementia. We aim to use NLP to detect signs of cognitive impairment from unstructured clinician notes by using deep learning techniques. Such a tool could also
be useful in recruiting into clinical trials as well as a
variety of research studies. We apply our deep learning algorithm to patients in Mass General Brigham (MGB) Healthcare who have genotype data available from the MGB BioBank. An overview of our project can be found in Appendix \ref{app:overview}. 

\section{Related Works}
\label{sec:RelatedWorks} 
Prior works have used NLP techniques to detect various diseases from EHR. \citep{rajkomar2018scalable} 
used recurrent neural networks (long short-term memory (LSTM)) among others to predict inpatient mortality using EHR data from the University of California, San Francisco (UCSF) from 2012 to 2016, and the University of Chicago Medicine (UCM) from 2009 to 2016.  \citep{glicksberg2018automated} 
performed phenotyping for diseases such as Attention Deficit Hyperactivity Disorder (ADHD) by clustering on word2vec embeddings from EHR of the Mount Sinai Hospital (MSH) in New York City. These studies have shown that the application of NLP techniques to EHR have improved disease detection, and that NLP techniques can be applied to dementia detecton to achieve similar results. Our work uses deep learning NLP techniques, which has achieved impressive results when applied to general text due to the use of word embeddings and attention-based models \citep{vaswani2017attention,mikolov2013distributed,pennington2014glove,peters2018deep, devlin2018bert}, but have had limited applications in healthcare, particularly in dementia research. 

\section{Dataset, Preprocessing, and Annotations}
\label{sec:Dataset+Preprocessing+Annotations}

\paragraph{Dataset}
\label{sec:Dataset} Our dataset consisted of a cohort (N = 16,428) of patients from the Mass General Brigham (MGB) HealthCare  (formerly Partner's Healthcare, comprising two major academic hospitals, community hospitals, and community health centers in the Boston area) system who were older than 60 years (as of July 13, 2021), had \textit{APOE} genotypes—the biggest genetic risk factor for Alzheimer's disease  \citep{mahley2000apolipoprotein}—available from the MGB BioBank, and at least one clinician note with a dementia-related keyword. \tableref{tab:table1} shows demographics of the cohort of patients.   

\paragraph{Preprocessing}
\label{sec:Preprocessing} For each patient in our dataset, we extracted unstructured clinician notes, identified matches to 18 dementia-related keywords (Appendix \ref{app:keywords}, including those related to memory, cognition, neuropsychological tests, and dementia diagnoses. We constructed sequences from the note text spanning each of these matches (of length 800 characters). Our cohort of 16,428 patients had 279,224 sequences with dementia-related keywords in total.

\begin{table}[htb]
\floatconts
{tab:table1}
{\caption{Demographics of Data}}
\centering
\resizebox{\columnwidth}{!}{
        \begin{tabular}{lccc}
            \toprule
            \bfseries Characteristic & \bfseries (N = 16428) \\
            \midrule
            \bfseries Age (years) mean (SD) & 73.01 (7.96) \\ 
            \bfseries Gender Male, \emph n (\%) & 8740 (53.2)\\ 
            \bfseries Race, \emph n (\%) \\ 
                \hspace{10mm} White & 14896 (90.7) \\
                \hspace{10mm} Other/Not Recorded & 608 (3.7) \\
                \hspace{10mm} Black & 570 (3.5) \\
                \hspace{10mm} Hispanic & 170 (1.0) \\
                \hspace{10mm} Asian & 168 (1.0) \\
                \hspace{10mm} Indigenous & 16 (0.01) \\
            \bfseries \textit{APOE} Genotype, \emph n (\%) \\ 
                \hspace{10mm} \textit{APOE} ${\bm{\varepsilon}}$2 & 2028 (12.3) \\
                \hspace{10mm} \textit{APOE} ${\bm{\varepsilon}}$3 & 10177 (62.0) \\
                \hspace{10mm} \textit{APOE} ${\bm{\varepsilon}}$4 & 4223 (25.7) \\
            \bfseries Average Speciality Visits (SD) & 1.67 (4.6) \\ 
            \bfseries Average PCP Encounters (SD) &  5.25 (5.63) \\ 
            \bottomrule
        \end{tabular}
}
\end{table}

\begin{figure}[h!] \label{app:examples}
\centering 
\includegraphics[scale = 0.35]{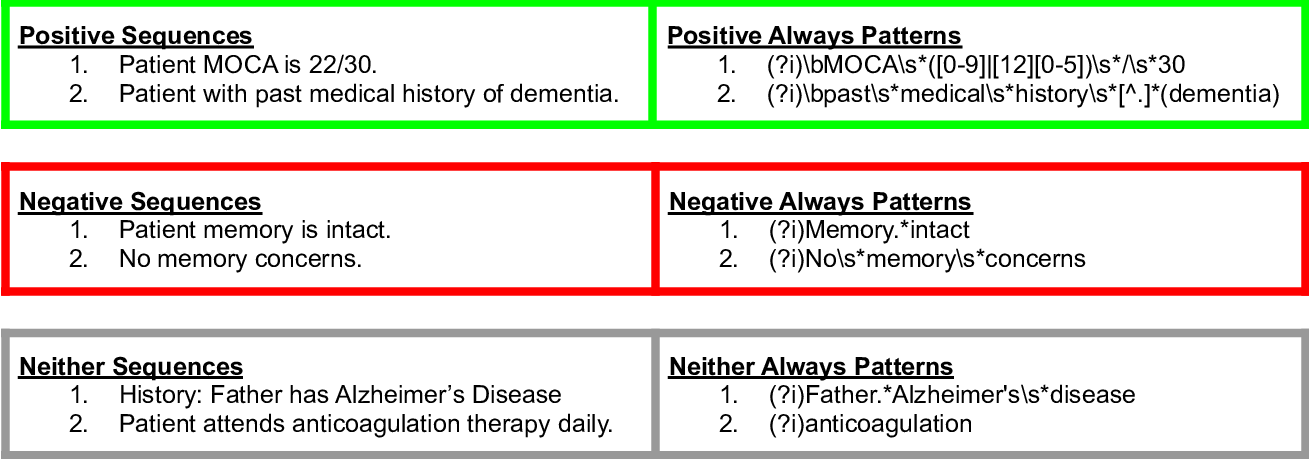}
\caption{Example Sequences}
\end{figure}

\paragraph{Annotations}
\label{sec:Annotations} A subset of sequences was annotated for indication of cognitive impairments. We defined cognitive impairment as evidence of MCI, where one cognitive domain is involved, or dementia, where more than one cognitive domain is involved and activities of daily living are affected. Concern from the family of the patient or the patient was not considered as cognitive impairment. Experts annotated sequences using a web-based annotation tool (Appendix \ref{app:slat}) and labeled them as 1) Yes, i.e., patient has CI; 2) No i.e., Patient does not have CI; and 3) Neither (NTR) i.e., sequence has no information on patient’s cognition. Appendix \ref{app:examples} shows example sequences for all 3 classes.

We assigned 5,000 diverse sequences containing at least one match to every keyword from 5,000 unique patients for labeling. In order to expedite annotations, we utilized an “always pattern" scheme. An always pattern is defined as a phrase or regex expression that in any context indicates the phrase will be labeled with a particular class (i.e. yes, no, or neither). Once an always pattern is defined, all other sequences that match the pattern are automatically labeled with that always pattern's class. For examples of always patterns for all three classes (Yes, No, Neither), see Appendix \ref{app:examples}. 

The final dataset of 8,656 annotated sequences from N = 2,487 unique patients was split between train (90\%) and holdout test (10\%) sets, stratified across label and proportion of sequences annotated manually and through always patterns. Validation datasets were split from the train set using techniques described in the Methodology section. 

\begin{table*}[hbtp]
\floatconts
{tab:table2}
{\caption{Model Performance}}
\centering
  {
  \begin{tabular}{lcccccccc}
    \toprule 
    \bfseries Model & \bfseries AUC & \bfseries Accuracy & \bfseries Sensitivity & \bfseries Specificity & \bfseries Micro F1  & \bfseries Macro F1 & \bfseries Weighted F1 \\ 
    \midrule
    TF-IDF & 0.95 & 0.84 & 0.83 & 0.92 & 0.84 & 0.81 & 0.84 \\
    ClincialBERT & 0.98 & 0.93 & 0.91 & 0.96 & 0.93 & 0.92 & 0.93 \\
    \bottomrule
  \end{tabular}
  }

  \label{tab:modelperf}
\end{table*}

\section{Methodology}

We developed two NLP models for the classification task and compared them to each other.

\label{sec:TFIDF}  
\paragraph{(1) Logistic Regression with TF-IDF Vectors} We performed TF-IDF (term frequency-inverse document frequency) vectorization on the annotated sequences and selected features based on a term's Pearson correlation coefficient with the cognitive impairment outcome. L1 Regularized logistic regression \citep{tibshirani1996regression} was applied with the annotated cognitive impairment labels. We used 10-fold cross validation to determine the optimal lambda value and correlation coefficient threshold to select features. 

\label{sec:Transformer}  
\paragraph{(2) Transformer Based Sequence Classification Language Model} We utilized a pre-trained language model called ClinicalBERT \citep{alsentzer-etal-2019-publicly}, which was trained on the MIMIC II \citep{saeed2011multiparameter} database containing EHR records from ICU patients. We used the implementation in the Huggingface Transformers \citep{Wolf2019HuggingFacesTS} and Simpletransformers \citep{simple2020thilina} packages. After text preprocessing, input texts were tokenized with the default tokenizer and converted to embeddings. The model was initialized with pre-trained parameters and later fine-tuned on our labeled training set. Optuna \citep{akiba2019optuna} was used to perform a 20-trial study and tune the learning rate, Adam epilson, and the number of train epochs on the held-out validation set to maximize AUC. An early stopping rule was used to prevent overfitting by ensuring that training stopped if the loss did not change substantially over 3 epochs.

\section{Results}
\label{sec:Results}  



We evaluated each model based on sequence level class assignments. Model performance for each model on the held-out test set are shown in Table \ref{tab:modelperf}. To compute each metric, we used the threshold that maximized accuracy. The TF-IDF model achieved an AUC of 0.95 and accuracy of 0.84. The 20 words with the highest correlation coefficients using TF-IDF word vectorization are shown in Appendix \ref{app:app3}. While TF-IDF was able to identify the presence of a keyword or always pattern in a sequence, it was unable to the leverage the context around each keyword match. The context of the keywords and the agents within the sentence often contained useful information regarding a patient's cognitive status. For example, the sentence "Patient is caregiver for wife who has dementia" has the keyword dementia, but does not pertain to the patient's cognitive diagnosis. This led the baseline TF-IDF model to incorrectly predict sequences as evidence of cognitive impairment, resulting in a large count of false positives.

\begin{figure}[htb]
\floatconts
{fig:umap}
{\caption{UMAP Clustering of ClinicalBert Embeddings}}
\centering 
\resizebox{\columnwidth}{!}{%

\includegraphics[width=.75\linewidth]{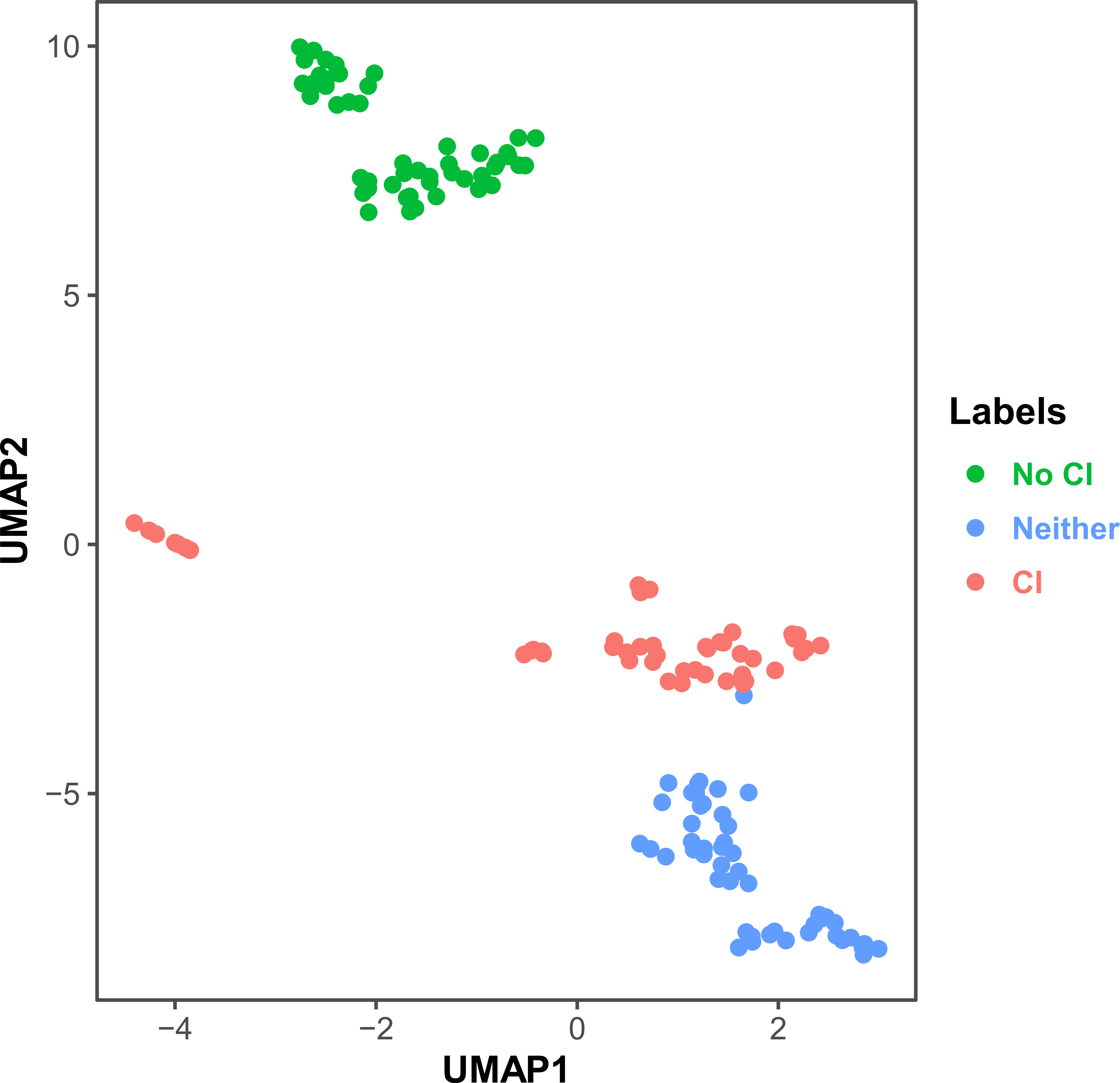}
}
\end{figure}

ClinicalBERT, with its more complex architecture, was able to leverage the context of the keyword matches within the sequences and overcome these issues. Indeed, using a small dataset of manually annotated sequences (n=150) which did not match an always-pattern, the ClinicalBERT embeddings were able to able accurately discriminate between all three classes (see Figure \ref{fig:umap}). The fine-tuned ClinicalBERT model achieved an AUC of 0.98 and substantially improved accuracy to 0.93 (specificity of 0.96, sensitivity of 0.91, micro F1 of 0.93, macro F1 of 0.92, and weighted F1 of 0.93). 

In order to generate patient level class assignments, we applied ClinicalBERT to all 186,730 sequences from the N = 13,941 unique patients that were not patient of our training/validation/set sets. With these sequence level predictions, we generated patient level class assignments by assigning patients a cognitive impairment label if their number of sequences predicted positive was greater than an empirically tuned threshold. We identified the most optimal threshold (from a range of 1 - 10) by comparing the percentage of patients being predicted as having cognitive impairment stratified by \textit{APOE} allele to the percentages of patients with cognitive impairment related Meds/ICD codes stratified by \textit{APOE} allele (M/I code column in \tableref{tab:bert}). \tableref{tab:bert} shows the comparison of Med/ICD codes to ClinicalBERT patient level class assignments with a sequence threshold of 2. As shown, ClinicalBERT was able to identify a significant proportion of patients that went undetected by current clinical methods, highlighting the utility of such a tool in a clinical setting. 

\begin{table}[htb]
\floatconts
{tab:bert}
{\caption{Comparison between Other Indicators of Cognitive Impairment and ClinicalBERT}}
\centering
\resizebox{\columnwidth}{!}{%
{\begin{tabular}{cccccc}

    \toprule 
    \bfseries  & \bfseries Count & \bfseries Yes (\%) & \bfseries No/Ntr (\%)  & \bfseries M/I (\%) \\ 
    \midrule
    \bfseries \textit{APOE} ${\bm{\varepsilon}}$2 & 1754 & 0.17 & 0.83 & 0.11  \\
    \bfseries \textit{APOE} ${\bm{\varepsilon}}$3 & 8751 & 0.17 & 0.83 & 0.11  \\
    \bfseries \textit{APOE} ${\bm{\varepsilon}}$4 & 3436 & 0.21 & 0.79 & 0.17  \\
    
    \bottomrule
  \end{tabular}
}
  
}

\bfseries M/I = Med/ICD Code 
\end{table}

\begin{figure}[htb]
\caption{Confusion Matrix Patient Level Prediction Counts for ClinicalBERT}
\resizebox{\columnwidth}{!}{%
\label{fig:fig2}
\centering 
\includegraphics[scale=0.5]{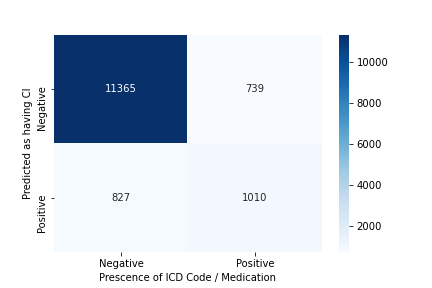}

}
\end{figure}

\section{Conclusion and Future Work} 
\label{sec:ConclusionFutureWork} We applied NLP algorithms to identify patients with cognitive impairment in EHR and compared a baseline TF-IDF model with an attention based deep learning model on performance of sequence level class assignment predictions. Our work can help address the underdiagnosis of dementia and alert primary care physcians to do a formal cognitive evaluation or refer to specialists. Such a tool can be used to generate cohorts for dementia research studies to identify risk and protective factors of dementia as well as recruit patients into observational studies or clinical trials. 

The deep learning model's performance was significantly better than the TF-IDF model as it was able to fully leverage the context of sequences. Our work illustrates the need of more complex, expressive language models for the nuanced task of detecting dementia in electronic health records.

We used the sequence level class assignments of the deep learning model to generate patient level classes. For a few patients, we manually verified that our model can successfully identify patients with cognitive impairment who lack dementia-related ICD codes or medications in their records. However, a lack of patient level annotations prevents us from measuring the true accuracy of these results. In order to address this issue, we plan to generate 1000 patient level class assignments using our annotation tool. We also plan to further improve the generalizability of our models by labeling more sequences that do not match an always pattern. An active learning loop will be used to pick particular patients and sequences by using entropy scores to label uncertain cases and UMAP clustering \citep{mcinnes2018umap} of ClinicalBERT word embeddings on the sequences of the N = 13,941 patients. The new gold-standard dataset will serve as the basis for the next iteration of the active learning loop to further improve model performance and detect patients with cognitive impairment with improved accuracy.

\clearpage


\bibliography{jmlr-sample}
\nocite{*}

\clearpage

\appendix

\section{Overview}
\begin{figure}[h] \label{app:overview}
\centering 
\includegraphics[scale = 0.60]{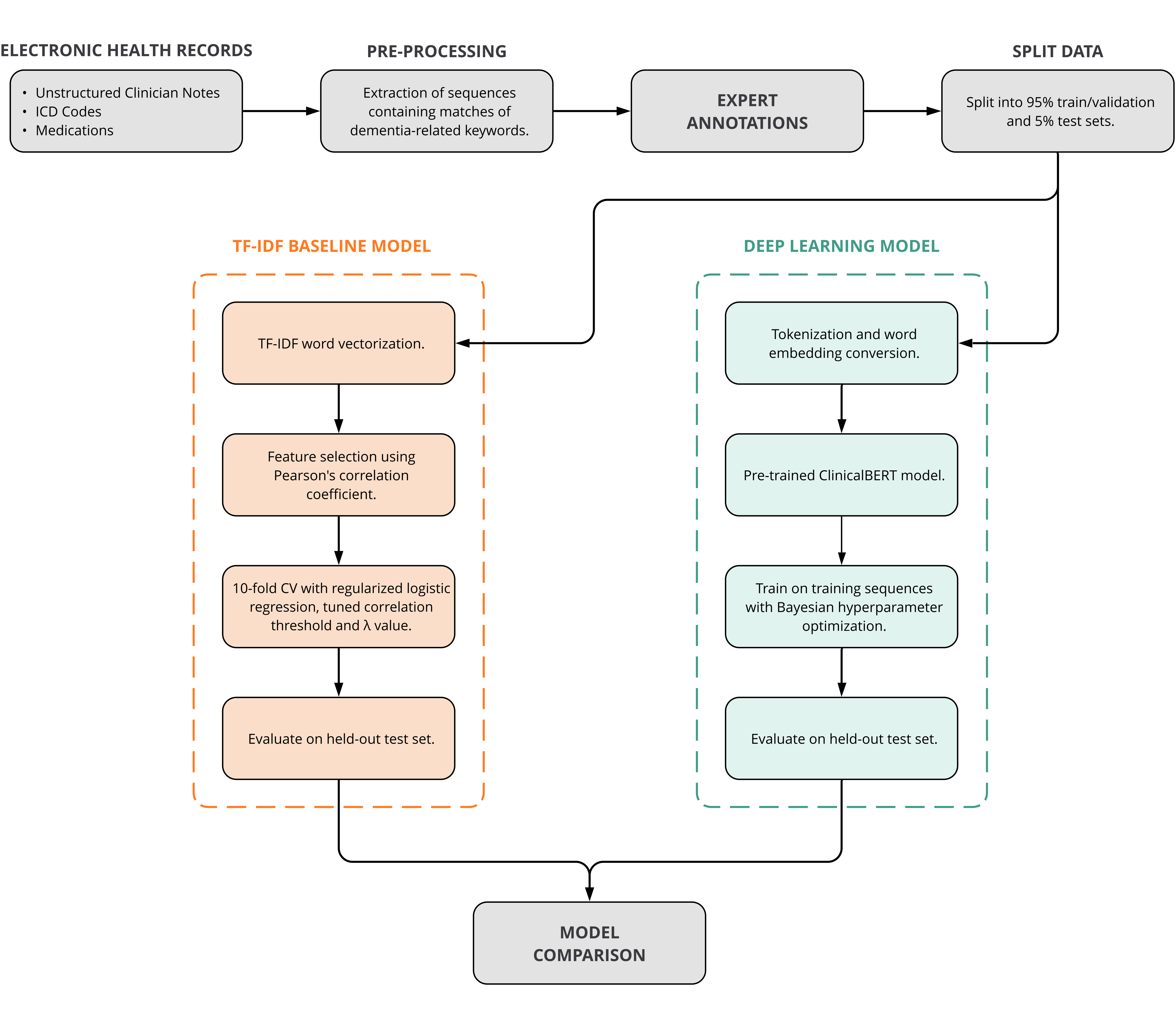}
\caption{Overview}

\end{figure}

\clearpage

\section{Pictures of UI Interface for Annotations} 
\label{app:slat}
\begin{figure}[h]
\centering 
\includegraphics[scale = 0.43]{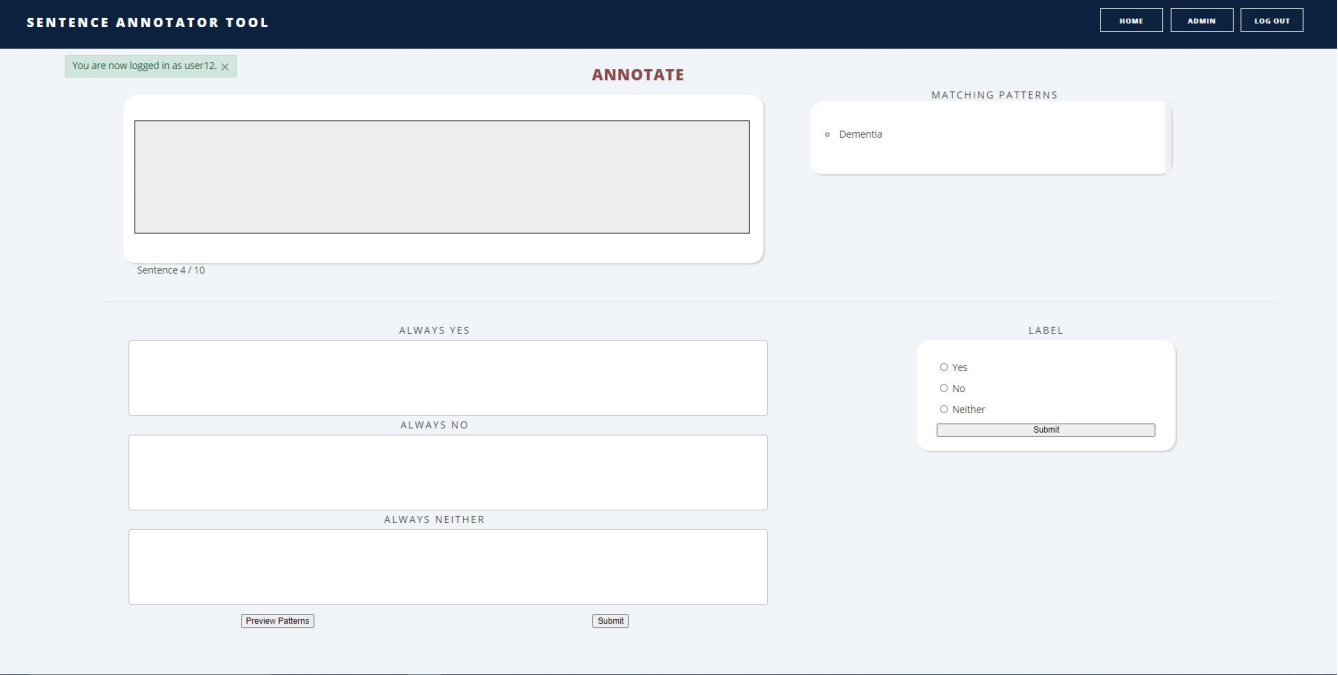}
\caption{Annotation UI}
\end{figure}

\begin{figure}[h]
\centering 
\includegraphics[scale = 0.3]{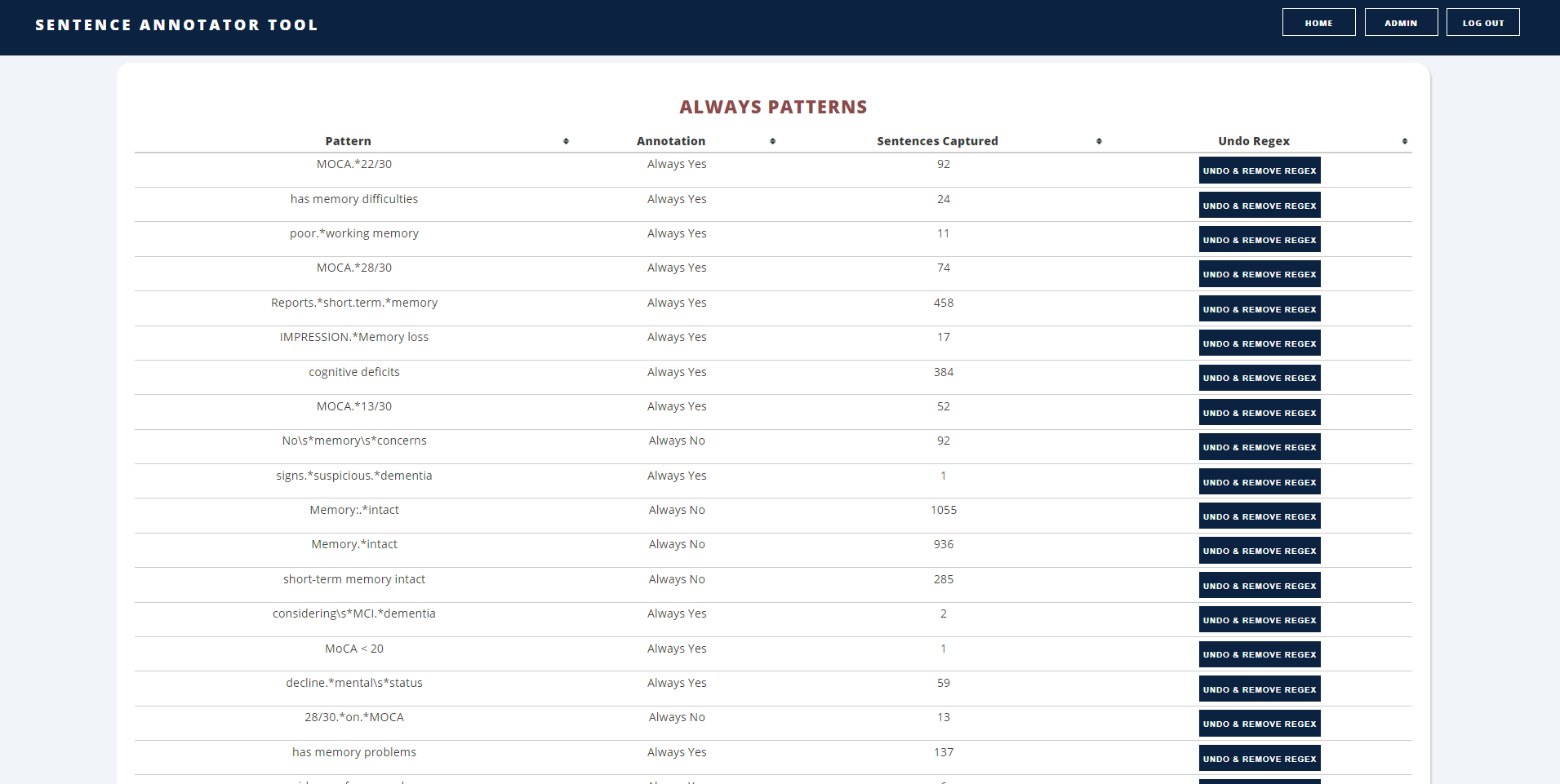}
\caption{Always Pattern List Generated by Annotations}
\end{figure}

\clearpage

\section{Example Sequences} 

\clearpage

\section{Keywords} 
\begin{table}[hbtp] 
\floatconts
{app:keywords}
{\caption{Keywords indicative of Cognitive Impairment}}
\centering
    \begin{tabular}{lcc}
        \toprule
        \bfseries Keyword & \bfseries Match Count \\
        \midrule
        
        \bfseries Memory & 109218 \\ 
        \bfseries Cognition & 87655 \\ 
        \bfseries Dementia & 51034 \\ 
        \bfseries Cerebral & 45886 \\ 
        \bfseries Cerebrovascular & 36370 \\ 
        \bfseries Cerebellar & 26863 \\
        \bfseries Cognitive Impairment & 20267 \\ 
        \bfseries Alzheimer & 20581 \\ 
        \bfseries MOCA & 9767 \\ 
        \bfseries Neurocognitive & 7711 \\ 
        \bfseries MCI & 3889 \\ 
        \bfseries Amnesia & 3695 \\ 
        \bfseries AD & 2673 \\ 
        \bfseries Lewy & 2561 \\ 
        \bfseries MMSE & 2134 \\ 
        \bfseries LBD & 224 \\ 
        \bfseries Corticobasal & 147 \\ 
        \bfseries Picks & 41 \\ 
        
        \bottomrule
        \end{tabular}
        
\end{table}

\section{Top TF-IDF Word Features} \label{app:app3}
\begin{table}[hbtp]
    \begin{tabular}{lcccc}
    \toprule 
    \bfseries Word & \bfseries Corr & \bfseries Word & \bfseries Corr \\
    \midrule
    
    \bfseries Intact &  0.56 & \bfseries Experiences & 0.36 \\
    \bfseries Oriented &  0.43 & \bfseries Associations & 0.36 \\ 
    \bfseries Concentration &  0.42 & \bfseries Homicidal & 0.36 \\ 
    \bfseries Orientation &  0.41 & \bfseries Observation & 0.36 \\
    \bfseries Sensorium &  0.40 & \bfseries Knowledge & 0.36 \\ 
    \bfseries Perceptions &  0.40 & \bfseries Abstract & 0.36 \\
    \bfseries Judgement &  0.39 & \bfseries Suicidal & 0.35 \\
    \bfseries Fund & 0.38 & \bfseries Attention & 0.35 \\
    \bfseries Insight &  0.36 & \bfseries Content & 0.34 \\ 
    \bfseries Ideation &  0.36 & \bfseries Thought & 0.34  \\  
    
    \bottomrule
  \end{tabular}
  {\caption{Top 20 TF-IDF Word Features and their Correlation Coefficient}}
\end{table}


    

\end{document}